\def\paperID{11652}
\def\confName{ICCV}
\def\confYear{2025}

\def\authorBlock{
Chunlin Wen\textsuperscript{1}, Yu Zhang\textsuperscript{1}\thanks{Corresponding author: zhang\_yu@seu.edu.cn}, Jie Fan\textsuperscript{2}, Hongyuan Zhu\textsuperscript{4}, Xiu-Shen Wei\textsuperscript{1},\\ Yijun Wang\textsuperscript{1}, Zhiqiang Kou\textsuperscript{1}, Shuzhou Sun \textsuperscript{3,5} \\
	\textsuperscript{1}School of Computer Science and Engineering, Southeast University
	\\
	\textsuperscript{2} Samsung Electronics (China) R\&D Centre   \quad
	\textsuperscript{3}Shanghai AI Laboratory
	\\
	\textsuperscript{4} Institute for Infocomm Research (I2R), A*STAR Singapore 138632 \\
	\textsuperscript{5}Center for Machine Vision and Signal Analysis (CMVS), University of Oulu
	 \\

}

\newif\ifreview 
\newif\ifarxiv 
\newif\ifcamera \newcommand{\cameraready}{\cameratrue}
\newif\ifrebuttal 

\cameraready
\pdfoutput=1
\documentclass[10pt,twocolumn,letterpaper]{article}
\ifreview \usepackage[review]{cvpr} \fi
\ifarxiv \usepackage[pagenumbers]{cvpr} \fi
\ifrebuttal \usepackage[rebuttal]{cvpr} \fi
\ifcamera \usepackage{cvpr} \fi
\usepackage{bbm}
\usepackage[accsupp]{axessibility}  

\usepackage{graphicx}	
\usepackage{amsmath}	
\usepackage{amssymb}	
\usepackage{booktabs}
\usepackage{times}
\usepackage{microtype}
\usepackage{epsfig}
\usepackage{caption}
\usepackage{float}
\usepackage{placeins}
\usepackage{color, colortbl}
\usepackage{stfloats}
\usepackage{enumitem}
\usepackage{tabularx}
\usepackage{xstring}
\usepackage{multirow}
\usepackage{xspace}
\usepackage{url}
\usepackage{subcaption}
\usepackage{xcolor}
\usepackage{tcolorbox}
\usepackage[hang,flushmargin]{footmisc}

\ifcamera \usepackage[accsupp]{axessibility} \fi

% ...

\ifarxiv  \fi

\newcommand{\R}[1]{{%
    \textbf{%
        \ifstrequal{#1}{1}{\textcolor{red}{R#1}}{%
        \ifstrequal{#1}{2}{\textcolor{blue}{R#1}}{%
        \ifstrequal{#1}{3}{\textcolor{magenta}{R#1}}{%
        \ifstrequal{#1}{4}{\textcolor{teal}{R#1}}{%
                           \textcolor{cyan}{R#1}%
        }}}}%
    }%
}}

\usepackage{xr-hyper}

\makeatletter
\newcommand*{\addFileDependency}[1]{
  \typeout{(#1)}
  \@addtofilelist{#1}
  \IfFileExists{#1}{}{\typeout{No file #1.}}
}

\makeatother
\newcommand*{\myexternaldocument}[1]{
    \externaldocument{#1}
    \addFileDependency{#1.tex}
    \addFileDependency{#1.aux}
}

\definecolor{cvprblue}{rgb}{0.21,0.49,0.74}
\usepackage[pagebackref,breaklinks,colorlinks,allcolors=cvprblue]{hyperref}
\usepackage[capitalize]{cleveref}
\crefname{section}{Sec.}{Secs.}
\crefname{table}{Table}{Tables}
\crefname{figure}{Fig.}{Figs.}
\ifarxiv \crefname{appendix}{App.}{Apps.}
\else \crefname{appendix}{Suppl.}{Suppls.} \fi
\unless\ifarxiv \myexternaldocument{_supplementary} \fi

\begin{document}
	\title{Object-level Correlation for Few-Shot Segmentation}
	\author{\authorBlock}
	\maketitle
	
	\begin{abstract}
\hspace{1em}Few-shot semantic segmentation (FSS) aims to segment objects of novel categories in the query images given only a few annotated support samples. Existing methods primarily build the image-level correlation between the support target object and the entire query image. However, this correlation contains the hard pixel noise, \textit{i.e.}, irrelevant background objects, that is intractable to trace and suppress, leading to the overfitting of the background. To address the limitation of this correlation, we imitate the biological vision process to identify novel objects in the object-level information. Target identification in the general objects is more valid than in the entire image, especially in the low-data regime. Inspired by this, we design an Object-level Correlation Network (OCNet) by establishing the object-level correlation between the support target object and query general objects, which is mainly composed of the General Object Mining Module (GOMM) and Correlation Construction Module (CCM). Specifically, GOMM constructs the query general object feature by learning saliency and high-level similarity cues, where the general objects include the irrelevant background objects and the target foreground object. Then, CCM establishes the object-level correlation by allocating the target prototypes to match the general object feature. The generated object-level correlation can mine the query target feature and suppress the hard pixel noise for the final prediction. Extensive experiments on PASCAL-${5}^{i}$ and COCO-${20}^{i}$ show that our model achieves the state-of-the-art performance.
\end{abstract}
	\vspace{-12pt}
\section{Introduction}
\vspace{-6pt}
\label{sec:intro}

\begin{figure}[tp]
\centering
{\includegraphics[width=3.3in]{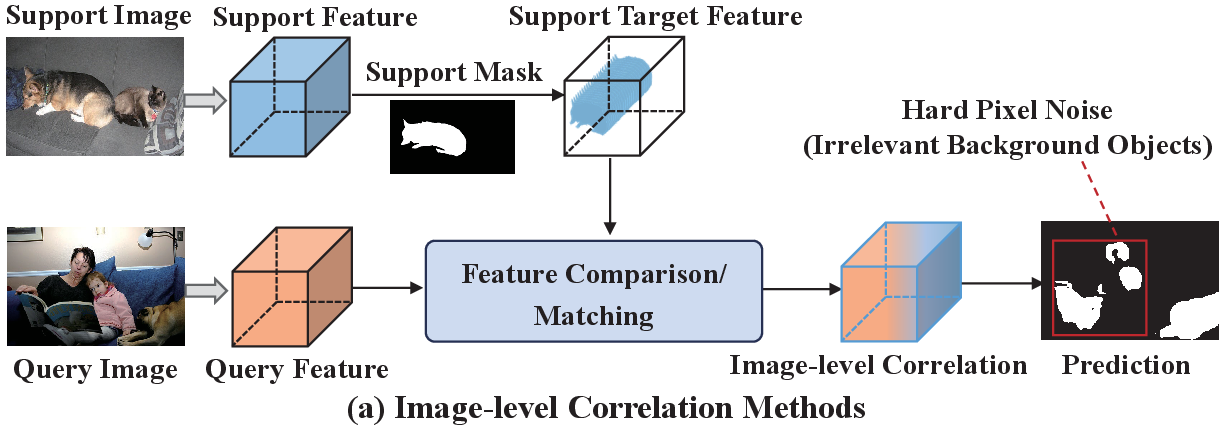}\vspace{0.5em}}
{\includegraphics[width=3.3in]{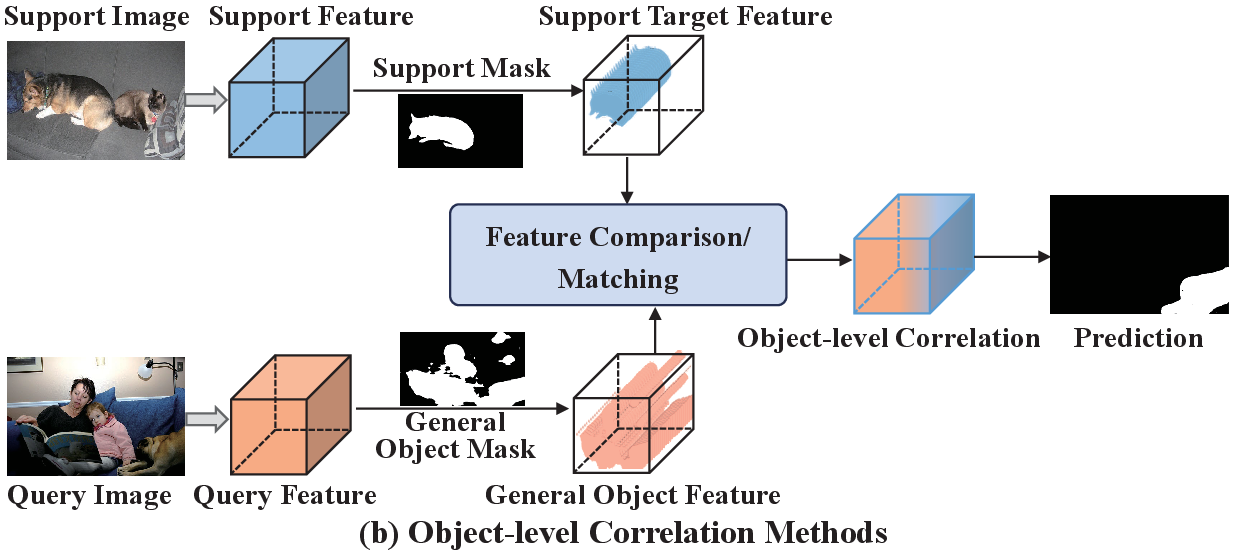}\vspace{-0.5em}}
\caption{Comparison between (a) previous image-level correlation method and (b) our object-level correlation method. (a) Previous image-level correlation methods focus on building the correlation between the support target feature and the entire query feature, leading to the hard pixel noise, such as, the real background object (books) and the irrelevant novel object (persons). (b) Our object-level correlation method is devoted to target selection from the general object feature by imitating the biological vision process. In this way, the generated correlation focuses on the target objects and suppresses the hard pixel noise.}
\vspace{-18pt}
\label{fig1}
\end{figure}

\hspace{1em}Semantic segmentation \cite{7298965,7803544,7913730,ronneberger2015u,Huang_2019_ICCV}, a fundamental task in computer vision \cite{simonyan2014very,10636800,10540001,he2016deep,ijcai2024p478}, has achieved significant progress in academia and industry. But these achievements primarily rely on large pixel-level annotated datasets, which demand extensive time and human effort. Moreover, semantic segmentation performance tends to be unsatisfactory when meeting the unseen novel classes. Under such circumstances, few-shot semantic segmentation (FSS) \cite{shaban2017one} is proposed by introducing few-shot learning \cite{snell2017prototypical, xu2023deep} into semantic segmentation. The FSS model aims to segment target novel objects in the original (query) image with a few reference (support) images. The key point for FSS is constructing the correlation between support and query information.

Previous traditional methods \cite{shaban2017one,dong2018few,tian2020prior,min2021hypercorrelation,zhang2019pyramid} focus on building the image-level correlation between the support target feature and the entire query feature by comparison and matching, as shown in Fig. \ref{fig1}(a). Based on the image-level correlation, they further segment the target object from the entire query feature through prototypical learning \cite{dong2018few,tian2020prior,zhang2020sg} or affinity learning \cite{zhang2019pyramid,min2021hypercorrelation,shi2022dense}. However, the image-level correlation generated by the entire query image tends to contain the hard pixel noise (\textit{i.e.}, irrelevant background objects) in the background. This noise mainly consists of real background objects, base objects, and irrelevant novel objects, representing the book, the sofa, and the person in the query image of Fig. \ref{fig1}, respectively. Therefore, as shown in the prediction of Fig. \ref{fig1}(a), this noise typically leads to inaccurate segmentation, like the book and the person. 
Some recent methods \cite{zhang2019canet,zhang2021few,Lang_2022_CVPR,liu2022learning,zhu2024addressing,xu2024eliminating} attempt to address these issues by eliminating the real background objects or base objects in the image-level correlation through post-processing.
Although suppressing most noise, these methods still ignore the elimination of irrelevant novel objects in the background. For example, in the query image of Fig. \ref{fig1}, the dog and person are both novel class objects, but only the dog is the target object that needs to be segmented. So, these methods fail to suppress the irrelevant novel object (person) in the background. 
All in all, existing methods face the following challenges:
1) Image-level correlation methods often incorporate real background objects, base objects, and irrelevant novel objects into the segmentation process, causing misclassification.
2) They still struggle to accurately identify target objects when multiple novel objects are present in the scene.

\vspace{-4.5pt}
To address the limitation of image-level correlation, we try to imitate the biological vision process: Saliency is computed in a pre-attentive manner across the entire visual field, and then the higher areas control target selection based on task-dependent cues \cite{itti2001computational}. In other words, the biological vision recognition system relies more on processing and understanding object-level information (saliency) than the entire image information. Target selection in the saliency (general objects) is more valid than in the entire image, especially in the low-data regime. Inspired by this, the salient information is first learned to construct the query general objects. Then, we identify the query target object from these objects with the guidance of the support target object (task-dependent cues). Following the above process, we propose an Object-level Correlation Network (OCNet) in Fig. \ref{fig1}(b), which establishes the object-level correlation between the support target feature and the query general object feature. Unlike the image-level correlation \cite{zhang2019canet,zhang2021few,Lang_2022_CVPR,liu2022learning,zhu2024addressing,xu2024eliminating}, the object-level correlation can accurately identify the target object in the foreground while suppressing the irrelevant objects in the background. Moreover, the support target more efficiently corresponds to the query target from the query general object than the entire query image \cite{shaban2017one,dong2018few,tian2020prior,min2021hypercorrelation,zhang2019pyramid}. Therefore, as shown in the prediction of Fig. \ref{fig1}(b), our model eliminates the hard pixel noise (\textit{i.e.}, the real background object (book) and irrelevant object (person)) in the background and segments the query target object (dog) correctly. 

Specifically, OCNet is mainly composed of the General Object Mining Module (GOMM) and Correlation Construction Module (CCM). Following the process of biological vision, we first propose the General Object Mining Module (GOMM) to generate the query general object feature. However, in the task of FSS, there are no given query masks to guide the learning of general objects. Therefore, we adopt the CAM \cite{zhou2016learning} to obtain the vanilla general object mask about the query image. Although identifying the most general objects, this vanilla mask sometimes does not contain total object information. To capture the lost information, we further integrate the high-level similarity mask into the vanilla mask and utilize the cross-attention \cite{NIPS2017_3f5ee243} to fuse the initial general object feature and the original query feature. After obtaining the query general object feature, we further establish the object-level correlation to identify the query target object based on the task-dependent cues (support target information). To this end, the Correlation Construction Module (CCM) is proposed that allocates the support frequency prototypes \cite{wen2024dual} to match the general object feature. Different from \cite{wen2024dual}, CCM introduces the prototype allocating mask to capture the target object by foreground prototypes and suppress the hard noise pixel by the background prototypes (ignored by \cite{wen2024dual}). Finally, our network can effectively segment the query target object from the object-level correlation. 

In summary, the contributions of this paper are concluded as follows:

\vspace{-2pt}
\begin{itemize}
\item{By imitating the biological vision process, we introduce the object-level correlation to address the limitation of image-level correlation, which refines the target object segmentation while suppressing the hard pixel noise.}
\item{We propose an Object-level Correlation Network (OCNet) that integrates general and high-level cues to generate the general object feature and further models the optimal allocating pattern to construct the object-level correlation.}
\item{Extensive experiments show that OCNet achieves state-of-the-art (SOTA) performance on few-shot segmentation.}
\end{itemize}

	\vspace{-5pt}
\section{Related Work}
\vspace{-5pt}
\label{sec:related}
\subsection{Semantic Segmentation}
\vspace{-4pt}
\hspace{1em}Semantic segmentation is a pivotal task in computer vision that aims to classify each pixel in an image according to a predefined set of semantic categories. Fully Convolutional Network (FCN) \cite{7298965} is the pioneering work in solving the problem of semantic segmentation, which replaces the fully connected layer in a classification framework with the convolution layer. Since then, tremendous progress has been made in this field, such as the encoder-decoder structure \cite{ronneberger2015u,7803544,xie2024squeeze,dong2025shape} for better feature extraction, dilated convolutions \cite{7913730,zhe2021dilated,Chen_2018_ECCV} to enlarge the receptive field, and pyramid pooling \cite{Zhao_2017_CVPR,Kirillov_2019_CVPR,lin2017refinenet} to aggregate multi-scale features. Moreover, some researchers \cite{Wang_2018_CVPR,fu2019dual,Huang_2019_ICCV,zhang2023attention,wu2024graph} focus on the efficient attention mechanism for capturing long-distance dependencies. However, the aforementioned methods rely heavily on extensive pixel-level annotations and exhibit limited generalization to novel classes under data-scarce conditions. This paper aims to tackle the above semantic segmentation limitation in the few-shot setting.

\vspace{-2pt}
\subsection{Few-Shot Semantic Segmentation}
\vspace{-4pt}
\hspace{1em}Few-shot semantic segmentation (FSS) \cite{shaban2017one} learns to generate dense predictions for novel class query images given the few pixel-wise annotated support images. Most existing FSS methods adopt the two-branch architecture, roughly divided into two categories: prototypical learning methods \cite{dong2018few,Lang_2022_CVPR,zhang2019canet,li2021adaptive,wen2024dual,liu2022dynamic,tian2020prior} and affinity learning methods \cite{peng2023hierarchical,zhang2021few,min2021hypercorrelation,zhang2019pyramid,shi2022dense,wang2020few,hong2022cost,xiong2022doubly}. Following PL \cite{dong2018few}, some prototypical learning methods extract the single global prototype \cite{zhang2019canet,Lang_2022_CVPR,tian2020prior} or multiple local prototypes \cite{li2021adaptive,liu2022dynamic,ijcai2022p143} from the support set to guide the query target segmentation. Notably, recent works focus on learn prototypes from other perspectives for further object information extraction, such as holistic prototypes \cite{cheng2022holistic}, self-support prototypes \cite{fan2022self}, frequency prototypes \cite{wen2024dual}, intermediate prototypes \cite{liu2022intermediate}, and so on. Besides, some affinity learning methods are proposed to preserve structural information lost by prototypical learning methods. They are devoted to building the dense pixel-level attention map between support and query images by graph attention mechanism \cite{zhang2019pyramid,wang2020few}, 4D convolutions \cite{min2021hypercorrelation,hong2022cost,xiong2022doubly}, Transformers \cite{peng2023hierarchical,zhang2021few,shi2022dense,wang2020few}, or Mamba \cite{xu2024hybrid}. However, previous works only emphasize the image-level correlation between support and query images, ignoring the object analysis and leading to the hard pixel noise in this correlation. Unlike them, our method learns all the object information from the query image, and further match the object information in object-level correlation to segment the target object.

\vspace{-2pt}
\section{Task Definition}
\vspace{-4pt}
\label{sec:task_def}
\hspace{1em}Following previous works \cite{dong2018few,Lang_2022_CVPR,peng2023hierarchical}, we adopt the standard few-shot semantic segmentation setting, \textit{i.e.}, episodic meta-training paradigm. Specifically, the dataset is divided into the training set ${D}_{train}$ and the testing set ${D}_{test}$. The ${C}_{train}$ (base) and ${C}_{test}$ (novel) object classes of two sets are disjoint (${C}_{train}\cap{C}_{test}=\varnothing$). Given a $K$-shot segmentation task, each episode consists of a query set ${Q}=\lbrace{I}_{q},{M}_{q}\rbrace$ and a support set ${S}=\lbrace{I}^{k}_{s},{M}^{k}_{s}\rbrace^{K}_{k=1}$, where ${I}\in\mathbb{R}^{{H}\times{W}\times{3}}$ and ${M}\in\lbrace0,1\rbrace^{{H}\times{W}}$ represent the input images and the corresponding binary masks, respectively. During training, the model segments the query object based on the $S$ and ${I}_{q}$ by iteratively sampling an episode from ${D}_{train}$. After that, the trained model is directly evaluated on the test episodes sampled from ${D}_{test}$ without further optimization. Note that both support masks ${M}_{s}$ and query masks ${M}_{q}$ are available during training, whereas only ${M}_{s}$ is accessible during testing.

	
\section{Method}
\label{sec: method}
\subsection{Overview}

\begin{figure*}[tp]
    \centering
    \includegraphics[width=\linewidth]{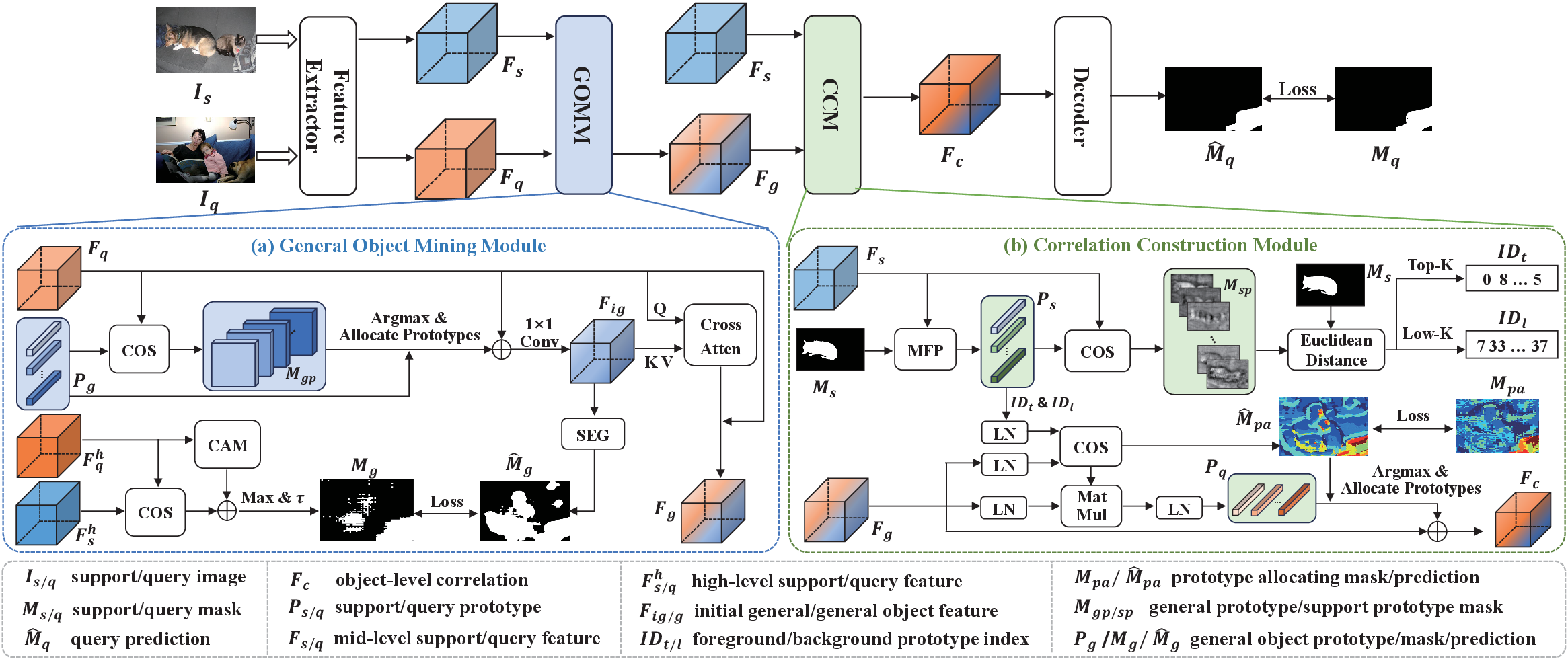}
    \vspace{-15pt}
    \caption{Overall architecture of our proposed OCNet. After extracting features from the pre-trained backbone, the General Object Mining Module (GOMM) (blue) first utilizes these features to capture the general object feature ${F}_{g}$ via learning the general object prototype. Then, the object-level correlation ${F}_{c}$ is constructed by support target information and ${F}_{g}$ in the proposed Correlation Construction Module (CCM) (green). Finally, ${F}_{c}$ is fed into the decoder for the final query prediction.}
     \vspace{-15pt}
    \label{fig2}
\end{figure*}

\textbf{Motivation.} \hspace{0.5em} Previous metric-based methods primarily construct the image-level correlation between support target objects and the entire query image by 4D convolutions \cite{min2021hypercorrelation}, Transformers \cite{peng2023hierarchical}, dense prototype comparison \cite{tian2020prior,Lang_2022_CVPR}, \textit{etc}. Unfortunately, it is difficult to directly associate the target object with the entire query image, and this correlation tends to be biased towards irrelevant background objects. This is because, compared to the non-object samples, the general objects are more similar to each other \cite{liang2023unknown}. Moreover, these methods regard irrelevant objects as background features during training. However, these objects are typically the novel target objects in the testing stage, which the model needs to predict. The excessive suppression of irrelevant objects is not conducive to segmenting novel objects. Therefore, the optimal pattern is to separate the general objects from the entire image and further identify the target object from the general objects by constructing the correlation between objects, like the process of biological vision \cite{itti2001computational}. Inspired by this, we propose the Object-level Correation Network (OCNet) to address the limitation of previous image-level correlation.
\\\textbf{Architecture.} \hspace{0.5em}As shown in Fig. \ref{fig2}, OCNet mainly consists of two major modules, \textit{i.e.}, the General Object Mining Module (GOMM) and the Correlation Construction Module (CCM). Specifically, we first follow the previous works \cite{tian2020prior,Lang_2022_CVPR,peng2023hierarchical} to extract the mid-level object feature ${F}_{s/q}$ and high-level object feature ${F}^{h}_{s/q}$ with the pre-trained backbone. Then, those features are delivered to GOMM to capture the general object feature ${F}_{g}$ from the query image through the general object prototype learning. After obtaining ${F}_{g}$, CCM further establishes the object-level correlation ${F}_{c}$ by allocating the support prototypes ${P}_{s}$ to correspond with ${F}_{g}$. Finally, the decoder can effectively predict results based on ${F}_{c}$. In this way, our model focuses on the correlation between support target object and query general objects rather than the entire query image. Compared to the previous image-level correlation, the object-level correlation ${F}_{c}$ can recognize the query target object while suppressing the hard pixel noise in the query general objects. 
\subsection{General Object Mining Module}
\hspace{1em} Following the biological vision pattern, we propose the General Object Mining Module (GOMM) to mine the general object features from the query images in Fig. \ref{fig2} (a). The general object mask ${M}_{g}$ is first generated to supervise the learning of the initial general object feature ${F}_{ig}$. Then, we further complement the information lost by ${F}_{ig}$ to obtain the general object feature ${F}_{g}$.
\\\textbf{General object mask.} \hspace{0.5em} Since there are no given query masks to guide the learning of general objects, we adopt CAM \cite{zhou2016learning} to obtain the vanilla general object mask of query image. However, the vanilla general object mask sometimes involves no target object information. To alleviate this problem, the prior query mask is integrated into the vanilla mask to generate the general object mask. Although this mask is not complete and precise, we only need the obscure location and enhancement of the general object, and the moderately uncompleted information favors the generalization and reconstruction ability of general object prototypes \cite{he2022masked,wang2024snida,hu2023suppressing}. Moreover, the cross-attention \cite{NIPS2017_3f5ee243} is utilized to alleviate the incompleteness. Specifically, given the high-level support feature ${F}^{h}_{s}\in\mathbb{R}^{{H}\times{W}\times{C}_{h}}$ and the high-level query feature ${F}^{h}_{q}\in\mathbb{R}^{{H}\times{W}\times{C}_{h}}$, we first use the CAM and the cosine similarity to obtain the vanilla general object mask and the prior query mask, respectively. Following PFENet\cite{tian2020prior}, we compute the pixel-wise cosine similarity ($HW \times HW$) between the query and support feature. For each query pixel, the maximum similarity across all support pixels is selected to generate the prior query mask ($HW \times 1$), which is reshaped to $H \times W\times1$ and normalized. Then, these masks are fused by computing the pixel-level maximum value. Finally, the mask threshold ${\tau}$ is used to segment the fused mask and generate the general object mask ${M}_{g}$:
\vspace{-5pt}
\begin{equation}
	\vspace{-20pt}
	\label{deqn_ex1}
	{M}_{g} = \mathbbm{1}_{\tau}{(Max({Cosine({F}^{h}_{q},{F}^{h}_{s})}\oplus{CAM({F}^{h}_{q})}))},
	\vspace{-0.5em}
\end{equation}
\begin{equation}
	\vspace{-0.5em}
\begin{split}
	\label{deqn_ex2}
	\mathbbm{1}_{\tau}(x)&=
	\begin{split}
		\begin{cases}
			1,&{x{\geq}{\tau}},\\
			0,&{{x{<}}{\tau}},
		\end{cases}
	\end{split}
\end{split}
\vspace{-4pt}
\end{equation}
where $Cosine(\cdot)$ denotes the cosine similarity, and $\oplus$ refers to the channel-wise concatenation. $\mathbbm{1}$ is the indicator function that adopts the mask threshold ${\tau}$ to control the general object sampling scope,  where ${\tau}$ is set to 0.6 in our experiment. Note that the background information in ${F}^{h}_{s}$ is filtered out by the support mask ${M}_{s}$.
\\\textbf{Initial general object feature.} \hspace{0.5em} After that, ${M}_{g}$ is utilized to guide the learning of the general object prototype ${P}_{g}$ and generate the initial general object feature ${F}_{ig}$. Specifically, we first randomly initialize the general object prototypes ${P}_{g}\in\mathbb{R}^{{N}_{g}\times{C}}$ and apply the cosine similarity to produce the general prototype masks ${M}_{gp}\in\mathbb{R}^{{H}\times{W}\times{N}_{g}}$ from the ${P}_{g}$ and query features ${F}_{q}$:
\vspace{-3pt}
\begin{equation}
	\vspace{-3pt}
	\label{deqn_ex3}
	{M}_{gp} = Cosine({F}_{q},{P}_{g}),
\end{equation}
where ${M}_{gp}$ indicates the pixel-level similarities between each  ${F}_{q}$ and ${P}_{g}$. 
After applying the argmax operation to ${M}_{gp}$, we obtain the guide map (${H \times W}$), where each pixel stores the index of its corresponding prototype in ${P}_{g}$. Using this map, we place the corresponding prototype at each position to generate the allocated prototypes (${H \times W \times C}$). Finally, we concatenate the allocated prototypes with the ${F}_{q}$ and adopt ${1}\times{1}$ convolution to reduce the channel number of the concatenated features, generating the initial general object feature ${F}_{ig}\in\mathbb{R}^{{H}\times{W}\times{C}}$:
\vspace{-3pt}
\begin{equation}
	\vspace{-3pt}
	\label{deqn_ex4}
	{F}_{ig} = {Conv}_{1\times1}(Alloc({P}_{g}, Argmax({M}_{gp}))\oplus{F}_{q}),
\end{equation}
where $Alloc(\cdot, Argmax({M}_{gp}))$ denotes the allocation based on $Argmax({M}_{gp})$. Later, the ${F}_{ig}$ is predicted by the segment head to generate the general object prediction ${\hat{M}}_{g}$.
\\\textbf{General object feature.} \hspace{0.5em} Since ${F}_{ig}$ is learned under the supervision of ${M}_{g}$, it is difficult for ${F}_{ig}$ to capture the entire general object information. Therefore, we utilize ${F}_{q}$ to complement the information lost by ${F}_{ig}$. Conversely, ${F}_{ig}$ further enhances the general object feature in ${F}_{q}$. Specifically, ${F}_{ig}$ is fused into ${F}_{q}$ by utilizing the cross-attention in a QKV manner:
\begin{equation}
	\label{deqn_ex5}
	{F}_{g} = Atten({F}_{q},{F}_{ig},{F}_{ig})+{F}_{q},
\end{equation}    
where $Atten(\cdot)$ denotes the cross-attention operator and ${F}_{g}\in\mathbb{R}^{{H}\times{W}\times{C}}$ is the general object feature.
\subsection{Correlation Construction Module}

\begin{figure}[tp]
	\centering
	\includegraphics[width=\linewidth]{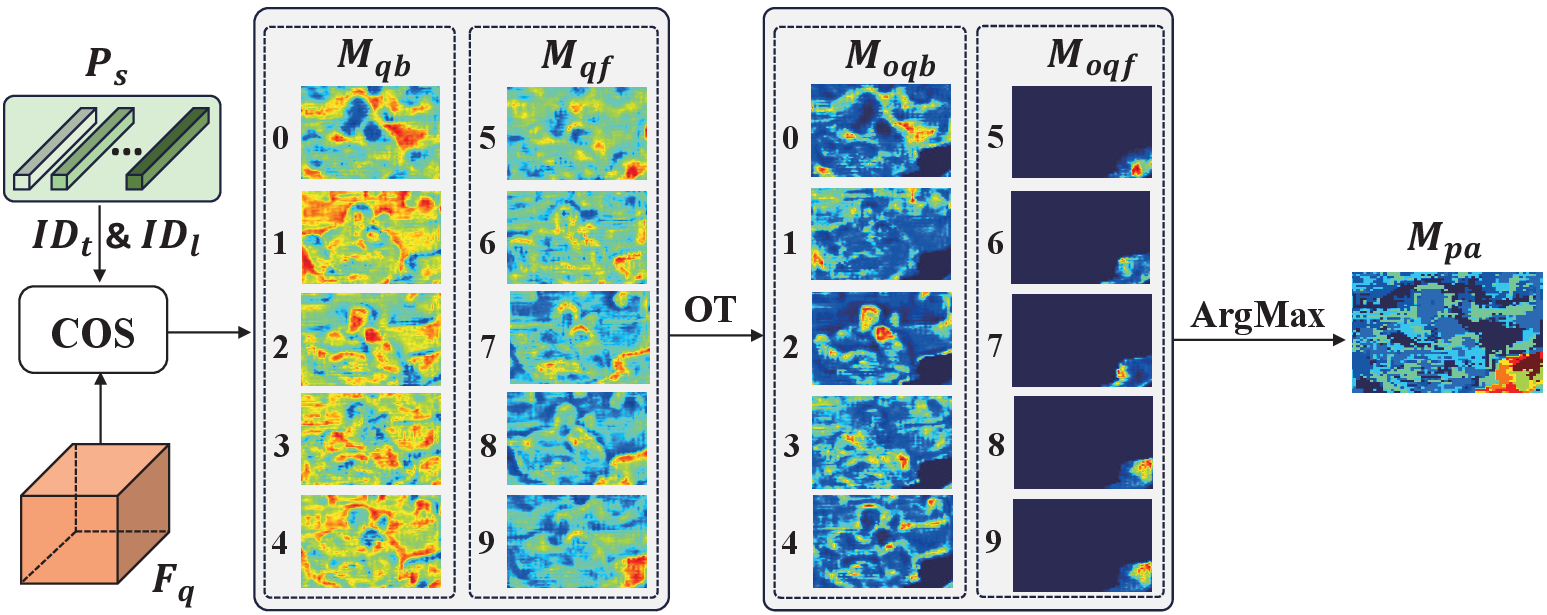}
	 \vspace{-15pt}
	\caption{Illustration of prototype allocating mask in CCM.}
	 \vspace{-15pt}
	\label{fig3}
\end{figure}

\hspace{1em}The Correlation Construction Module (CCM) aims to construct the object-level correlation between the support target object and the query general objects, as shown in Fig. \ref{fig2} (b). To this end, there are two key points: (i) gaining rich and complete support information; (ii) modeling optimal construction pattern. The details are as follows. 
\\\textbf{Support prototypes.} \hspace{0.5em} To gain rich and complete support information, we adopt multi-frequency pooling \cite{wen2024dual} to generate the support prototype ${P}_{s}$. However, in \cite{wen2024dual}, they only utilize the target object information in the support foreground prototypes, but ignore the rich and practical background prototypes. To alleviate this limitation, we not only uses the foreground prototypes to capture the global and local target object information, but also applies the background prototypes to suppress the hard pixel noise.

Specifically, given the support features ${F}_{s}$ and the support mask ${M}_{s}$, we first adopt multi-frequency pooling \cite{wen2024dual} to generate the support prototype ${P}_{s}\in\mathbb{R}^{{L}\times{C}}$ in the frequency domain (as illustrated in \cite{wen2024dual}, $L=49$):  
\vspace{-3pt}
\begin{equation}
	\vspace{-3pt}
	\label{deqn_ex6}
	{P}_{s} = MFP({F}_{s}, {M}_{s}),
\end{equation}
where $MFP(\cdot)$ denotes multi-frequency pooling. Then, the cosine similarity activates ${P}_{s}$ to generate all the support prototype mask ${M}_{sp}\in\mathbb{R}^{{H}\times{W}\times{L}}$:
\vspace{-3pt}
\begin{equation}
	\vspace{-3pt}
	\label{deqn_ex7}
	{M}_{sp} = Cosine({F}_{s}, {P}_{s}),
\end{equation}
More efficient prototypes tend to generate masks that exhibit higher similarity to ${M}_{s}$. In other words, the foreground prototypes strengthen the target objects, but background prototypes suppress the target objects. Therefore, we apply the euclidean distance between ${M}_{sp}$ and ${M}_{s}$ to select the foreground prototype index ${ID}_{t}\in\mathbb{R}^{{N}_{s}}$ and the background prototype index ${ID}_{l}\in\mathbb{R}^{{N}_{s}}$:  
\vspace{-5pt}
\begin{equation}
	\vspace{-10pt}
	\label{deqn_ex8}
	{ID}_{t} = Topk(Dis({M}_{sp}, {M}_{s}), {N}_{s}),
	\vspace{-1em}
\end{equation}
\begin{equation}
	\label{deqn_ex9}
	{ID}_{l} = Lowk(Dis({M}_{sp}, {M}_{s}), {N}_{s}),
	\vspace{-1pt}
\end{equation}
where the $Dis(\cdot)$ denotes the euclidean distance. $Topk(\cdot)$ and $Lowk(\cdot)$ denote the indices corresponding to the ${N}_{s}$ largest and smallest similarity scores, respectively.
\\\textbf{Prototype allocating mask.} \hspace{0.5em} When ${ID}_{t}$ and ${ID}_{l}$ are obtained, the corresponding prototypes are selected from ${P}_{s}$, which contain rich and complete support information. However, it is intractable to directly allocate the selected prototypes to construct the correlation with the general object feature ${F}_{g}$. To make the selected prototypes content-aware, we model the prototype allocation as the Optimal Transport (OT) problem, as shown in Fig. \ref{fig3}. Using the Sinkhorn algorithm with entropic regularization \cite{cuturi2013sinkhorn}, OT finds the transportation plan (i.e., the optimal query masks) with minimal global transportation cost to supervise prototype allocation.
  
Specifically, we first choose the corresponding prototypes, and produce the query foreground masks ${M}_{qf}\in\mathbb{R}^{{H}\times{W}\times{{N}_{s}}}$ by computing the cosine similarity between the selected prototypes and the query features ${F}_{q}$:
\vspace{-3pt}
\begin{equation}
	\vspace{-3pt}
	\label{deqn_ex10}
	{M}_{qf} = Cosine({F}_{q},Select({P}_{s}, {ID}_{t})).
\end{equation}
where $Select(\cdot, {ID}_{t})$ means the selection based on ${ID}_{t}$.
Then, the background pixels in ${M}_{qf}$ are filtered out by ${M}_{q}$, and the size of ${M}_{qf}$ is reshaped to ${N}_{f}\times{{N}_{s}}$, where ${N}_{f}$ is the amount of the query foreground pixels. After that, the cost matrix and the transport matrix are formulated as $(1-{M}_{qf})\in\mathbb{R}^{{{N}_{f}}\times{{N}_{s}}}$ and $T\in\mathbb{R}^{{{N}_{f}}\times{{N}_{s}}}$, where the lower transport cost corresponds to the higher similarity in ${M}_{qf}$. We define the optimization objective as: 
\vspace{-3pt}
\begin{equation}
	\vspace{-3pt}
	\label{deqn_ex11}
	\begin{split}
	\underset{T\in\tau}{\min}{\mathcal{L}}_{ot} = &Tr({T}^\top(1-{M}_{qf}))+\epsilon{H}(T),
	\\[-0.5em] s.t. &\quad T1=\mu, {T}^\top1=\nu
	\end{split}
\end{equation}
where ${H}(T)=-\sum_{ij}{T}_{ij}{\log}{{T}_{ij}}$ is the entropy regularizer, and $\epsilon$ is empirically set to 0.05 for controlling the smoothness of the transport matrix $T$. The transport matrix $T$ is constrained by $\mu = \frac{1}{{N}_{f}}{1}$ and $\nu = \frac{1}{{N}_{s}}{1}$. After performing several Sinkhorn iterations \cite{cuturi2013sinkhorn} to optimize Eq.~\ref{deqn_ex11}, the optimal transportation matrix ${T}^{*}$ is efficiently obtained and subsequently zero-padded at the corresponding background positions. By reshaping ${T}^{*}$ to ${{H}\times{W}\times{{N}_{s}}}$, the optimal query foreground mask ${M}_{oqf}$ is produced. Given the background frequency prototype index ${ID}_{l}$, we follow the same process of ${M}_{oqf}$ to generate the optimal query background mask ${M}_{oqb}$. Finally, the prototype allocating mask ${M}_{pa}\in\mathbb{R}^{{H}\times{W}\times{1}}$ is derived from the most optimal prototype indexes, which are selected from the optimal query masks by the argmax operator:
\vspace{-3pt}
\begin{equation}
	\vspace{-3pt}
	\label{deqn_ex12}
	{M}_{pa} = Argmax({M}_{oqb}\oplus{M}_{oqf}).
\end{equation}
\vspace{-1.5em}
\\\textbf{Correlation construction.} \hspace{0.5em} After obtaining ${M}_{pa}$, we can model optimal construction pattern. ${M}_{pa}$ supervises prototype allocation and guides the interaction between support and query features to derive the object-level correlation ${F}_{c}$. In this way, ${F}_{c}$ not only discriminates the target query objects, but also suppresses the hard pixel noise. Specifically, the prototype allocating prediction ${\hat{M}_{pa}}\in\mathbb{R}^{{H}\times{W}\times{2{N}_{s}}}$ is first captured by the cosine similarity:
\vspace{-3pt}
\begin{equation}
	\vspace{-3pt}
	\label{deqn_ex13}
	\hat{M}_{pa} = Cosine(LN({F}_{g}), LN(Select({P}_{s}, {ID}_{l}\&{ID}_{t}))),
\end{equation}
where $LN(\cdot)$ means the linear layers. Then, we integrate $\hat{M}_{pa}$ into ${F}_{g}$ through matrix multiplication, resulting in the query prototype ${P}_{q}\in\mathbb{R}^{{2{N}_{s}}\times{C}}$: 
\vspace{-3pt}
\begin{equation}
	\vspace{-3pt}
	\label{deqn_ex14}
	{P}_{q} = {LN}(MatMul(\hat{M}_{pa},{LN}({F}_{g}))).
\end{equation}
Owing to fusing the support and query information, ${P}_{q}$ can bridge the gap between the support and query sets. Given ${P}_{q}$ and $\hat{M}_{pa}$, we follow the same allocation operator as the Eq. \ref{deqn_ex4} and concatenate the ${F}_{g}$ to construct the object-level correlation ${F}_{c}\in\mathbb{R}^{{H}\times{W}\times{2C}}$: 
\vspace{-3pt}
\begin{equation}
	\vspace{-3pt}
	\label{deqn_ex15}
	{F}_{c} = Alloc({P}_{q}, Argmax(\hat{M}_{pa}))\oplus{F}_{g}.
\end{equation}

Finally, ${F}_{c}$ is passed into the decoder to obtain the query prediction $\hat{M}_{q}$, where the decoder employs FPN structure \cite{lin2017feature} to capture multi-scale object information and refine the final segmentation result.

\subsection{Training Loss}
\hspace{1em} We apply the cross entropy loss to supervise the learning of the query prediction $\hat{M}_{q}$, the general object prediction ${\hat{M}}_{g}$, and the prototype allocating prediction ${\hat{M}_{pa}}$. Therefore, the final training loss includes three parts: the target segmentation loss ${\mathcal{L}}_{t}$, the general segmentation loss ${\mathcal{L}}_{g}$, and the prototype allocation loss ${\mathcal{L}}_{p}$:
\begin{equation}
 	\label{deqn_ex16}
 	{{\mathcal{L}}_{f}} = {{\mathcal{L}}_{t}}+{{\mathcal{L}}_{g}}+{{\mathcal{L}}_{p}},
\end{equation} 
where ${\mathcal{L}}_{t}=CE(\hat{M}_{q},{M}_{q})$, ${\mathcal{L}}_{g}=CE({\hat{M}}_{g},{M}_{g})$, and ${\mathcal{L}}_{p}=CE({\hat{M}_{pa}},{M}_{pa})$.  With the supervision of ${M}_{g}$ and ${M}_{pa}$, our GOMM and CCM can effectively learn the general objects and construct the object-level correlation, respectively.

\begin{table*}
	\belowrulesep=0pt
	\aboverulesep=0pt
	\centering \scriptsize
	\caption{mIoU and FB-IoU performance of 1-shot and 5-shot segmentation on PASCAL-${5}^{i}$. The best performances are highlighted in bold.}
	\vspace{-5pt}
	\setlength{\tabcolsep}{4pt}
	\renewcommand{\arraystretch}{1.4}
	\begin{tabular}{l|c|c|cccc|c|c|cccc|c|c}
		\toprule
		\multirow{2}[1]{*}{Method} & \multirow{2}[1]{*}{Input Resolution} & \multirow{2}[1]{*}{Backbone} & \multicolumn{6}{c|}{1-shot} & \multicolumn{6}{c}{5-shot} \\
		\cmidrule{4-15}
		& & & Fold0 & Fold1 & Fold2 & Fold3 & Mean & FB-IoU & Fold0 & Fold1 & Fold2 & Fold3 & Mean & FB-IoU \\
		\cmidrule{1-15}
		PANet \cite{wang2019panet}(ICCV’19) &  $417 \times 417$ & \multicolumn{1}{c|}{\multirow{7}[2]{*}{VGG-16}} & 42.3 & 58.0 & 51.1 & 41.2 & 48.1 &66.5 & 51.8 & 64.6 & 59.8 & 46.5 & 55.7 &70.7\\
		PFENet \cite{tian2020prior}(TPAMI’20) &  $473 \times 473$ &  & 56.9 & 68.2 & 54.4 & 52.4 & 58.0 & 72.0 & 59.0 & 69.1 & 54.8 & 52.9 & 59.0 & 72.3 \\
		BAM \cite{Lang_2022_CVPR}(CVPR’22) & $473 \times 473$ &  & 63.2 & 70.8 & 66.1 & 57.5 & 64.4 & 77.3 & 67.4 & 73.1 & 70.6 & 64.0 & 68.8 & 81.1 \\
		HDMNet \cite{peng2023hierarchical}(CVPR’23) & $473 \times 473$  &  & 64.8 & 71.4 & 67.7 & 56.4 & 65.1 & - & 68.1 & 73.1& 71.8 & 64.0 & 69.3 & - \\
		AENet \cite{xu2024eliminating}(ECCV’24) & $473 \times 473$ &  & 66.3 & 73.3 & 68.5 & 58.4 & 66.6 & 79.0 & 70.8 & 75.1 & 72.2 & 64.2 & 70.6 & 81.8 \\
		HMNet \cite{xu2024hybrid}(NIPS’24) & $473 \times 473$ &  & 66.7 & \textbf{74.5} & \textbf{68.9} & 59.0 & 67.3 & 79.2 & 70.5 & \textbf{76.0} & 72.2 & 65.7 & 71.1 & \textbf{82.6} \\
		\cmidrule{1-2} \cmidrule{4-15}
		OCNet (ours) & $473 \times 473$ &  & \textbf{69.3} & 74.1 & 68.7 & \textbf{60.7} & \textbf{68.2} & \textbf{80.3} & \textbf{72.0} & 75.6 & \textbf{72.6} & \textbf{67.4} & \textbf{71.9} & \textbf{82.6} \\
		\cmidrule{1-15}
		
		CANet \cite{zhang2019canet}(CVPR’19) & $321\times 321$  &\multicolumn{1}{c|}{\multirow{7}[2]{*}{ResNet-50}}  
		&52.5 &65.9 &51.3 &51.9 &55.4 &66.2 &55.5 &67.8 &51.9 &53.2 &57.1 &69.6\\
		
		PFENet \cite{tian2020prior}(TPAMI’20) & $473 \times 473$  & & 61.7 & 69.5 & 55.4 & 56.3 & 60.8 & 73.3 & 63.1 & 70.7 & 55.8 & 57.9 & 61.9 & 73.9 \\
		BAM \cite{Lang_2022_CVPR}(CVPR’22) & $473 \times 473$ &  & 69.0 & 73.6 & 67.6 & 61.1 & 67.8 & 79.7 & 70.6 & 75.1 & 70.8 & 67.2 & 70.9 & 82.2 \\
		AENet \cite{xu2024eliminating}(ECCV’24) & $473 \times 473$ &  & 72.2 & 75.5 & 68.5 & 63.1 & 69.8 & 80.8 & 74.2 & 76.5 & 74.8 & 70.6 & 74.1 & 84.5 \\
		HMNet \cite{xu2024hybrid}(NIPS’24) & $473 \times 473$ &  & 72.2 & 75.4 & 70.0&  63.9 & 70.4 & 81.6 & 74.2 & 77.3 & \textbf{74.1} & \textbf{70.9} & 74.1 & 84.4\\
		
		ABCB \cite{zhu2024addressing}(CVPR’24) & $473 \times 473$ &  & 72.9 & \textbf{76.0} & 69.5 & 64.0 & 70.6 & - & 74.4 & \textbf{78.0} & 73.9 & 68.3 & 73.6 & - \\
		\cmidrule{1-2} \cmidrule{4-15}
		OCNet (ours) & $473 \times 473$ &  & \textbf{73.5} & 75.9 & \textbf{71.1} & \textbf{64.9} & \textbf{71.4} & \textbf{82.2} & \textbf{75.9} & 77.1 & \textbf{74.1} & \textbf{70.9} & \textbf{74.5} & \textbf{84.7} \\
		
		\bottomrule
	\end{tabular}
	\vspace{-8pt}
	\label{tab1}
\end{table*}

\section{Experiments}
\label{sec:experiments}
\subsection{Experimental Settings}
\textbf{Datasets.} \hspace{0.5em} Our model is evaluated on two widely-used benchmark datasets: PASCAL-$5^{i}$ \cite{shaban2017one} and COCO-$20^{i}$ \cite{nguyen2019feature}. PASCAL-$5^{i}$ is built from PASCAL VOC 2012 \cite{everingham2010pascal} with additional annotations from SDS \cite{hariharan2011semantic}, while COCO-$20^{i}$ is constructed based on MSCOCO dataset \cite{lin2014microsoft}. To be consistent with previous works \cite{shaban2017one, nguyen2019feature, tian2020prior, Lang_2022_CVPR, peng2023hierarchical}, we adpot the cross-validation manner. Specifically, the total categories are partitioned into 4 folds, where the each fold consists of 5 and 20 classes for PASCAL-$5^{i}$ and COCO-$20^{i}$, respectively. Then, we train the model on three folds, while using the remaining one fold for testing.  During meta-testing, 1,000 episodes are sampled from the test set for evaluating.
\\\textbf{Implementation Details.} \hspace{0.5em}In our experiment, two different backbone networks (VGG-16 \cite{simonyan2014very} and ResNet-50 \cite{he2016deep}) are chosen as the feature extractor to extract the mid-level and high-level features. Following \cite{tian2020prior,Lang_2022_CVPR,peng2023hierarchical}, these backbones are pre-trained on ImageNet \cite{russakovsky2015imagenet} and freezing parameters during all stages. Meanwhile, we apply the query generalization strategy \cite{wen2024dual} to fuse more target object semantic information into the general object feature. The model is trained with the SGD optimizer on PASCAL-$5^{i}$ for 200 epochs and COCO-$20^{i}$ for 75 epochs, where the learning rate and batch size are 0.005 and 4, respectively. Moreover, we employ the same data augmentation setting as \cite{tian2020prior} and crop images to the size 473 × 473 for PASCAL-$5^{i}$ and 641 × 641 for COCO-$20^{i}$ for training.  For $K$-shot setting, we average the support prototypes following \cite{tian2020prior, wen2024dual}. Our model is built upon the Pytorch framework and all experiments are conducted on the NVIDIA GeForce RTX 3090 GPUs.
\\\textbf{Evaluation Metrics.} \hspace{0.5em}Following common baselines \cite{tian2020prior,Lang_2022_CVPR,peng2023hierarchical}, mean intersection over union (mIoU) and foreground-background IoU (FB-IoU) are adopted as the evaluation metrics for experiments.
\begin{table*}
	\belowrulesep=0pt
	\aboverulesep=0pt
	\centering \scriptsize
	\caption{mIoU and FB-IoU performance of 1-shot and 5-shot segmentation on COCO-${20}^{i}$.  The best performances are highlighted in bold.}
	\vspace{-6pt}
	\setlength{\tabcolsep}{4pt}
	\renewcommand{\arraystretch}{1.45}
	\begin{tabular}{l|c|c|cccc|c|c|cccc|c|c}
		\toprule
		\multirow{2}[1]{*}{Method} & \multirow{2}[1]{*}{Input Resolution} & \multirow{2}[1]{*}{Backbone} & \multicolumn{6}{c|}{1-shot} & \multicolumn{6}{c}{5-shot} \\
		\cmidrule{4-15}
		& & & Fold0 & Fold1 & Fold2 & Fold3 & Mean & FB-IoU & Fold0 & Fold1 & Fold2 & Fold3 & Mean & FB-IoU \\
		\cmidrule{1-15}
		
		FWB \cite{nguyen2019feature}(ICCV’19) &  $512 \times 512$ & \multicolumn{1}{c|}{\multirow{7}[3]{*}{VGG-16}} & 18.4  & 16.7  & 19.6  & 25.4  & 20.0  & -  & 20.9  & 19.2  & 21.9  & 28.4  & 22.6  & - \\
		PFENet \cite{tian2020prior}(TPAMI’20) &  $641 \times 641$ &       & 35.4  & 38.1  & 36.8  & 34.7  & 36.3  & 63.3  & 38.2  & 42.5  & 41.8  & 38.9  & 40.4  & 65.0 \\
		BAM \cite{Lang_2022_CVPR}(CVPR’22) &  $641 \times 641$ &       & 39.0  & 47.0  & 46.4  & 41.6  & 43.5  & -  & 47.0  & 52.6  & 48.6  & 49.1  & 49.3  & - \\
		HDMNet \cite{peng2023hierarchical}(CVPR’23)&  $633 \times 633$  &       & 40.7  & 50.6  & 48.2  & 44.0  & 45.9  & -  & 47.0  & 56.5  & 54.1  & 51.9  & 52.4  & - \\
		SCCAN \cite{xu2023self}(ICCV’23)&  $473 \times 473$  &       & 38.3  & 46.5  & 43.0  & 41.5  & 42.3  & 66.9  & 43.4  & 52.5  & 54.5  & 47.3  & 49.4  & 71.8 \\
		AENet \cite{xu2024eliminating}(ECCV’24)&  $473 \times 473$  &       & 40.3  & 50.4  & 47.9 & 44.9  & 45.9  & 71.2 & 45.8 & 56.3 & \textbf{55.8} & \textbf{53.4} & 52.8 & 74.3 \\
		\cmidrule{1-2}\cmidrule{4-15}    
		OCNet(ours) &  $641 \times 641$  & & \textbf{42.4} & \textbf{51.3}  & \textbf{48.5} & \textbf{45.4} & \textbf{46.9} & \textbf{71.5} & \textbf{47.3}  & \textbf{57.3}  & 55.0 & 52.7 & \textbf{53.1} & \textbf{76.2} \\
		\cmidrule{1-15}

		PFENet \cite{tian2020prior}(TPAMI’20)&  $641 \times 641$  &\multicolumn{1}{c|}{\multirow{8}[2]{*}{ResNet-50}}& 36.5  & 38.6  & 35.0  & 33.8  & 35.8  & -  & 36.5  & 43.3  & 38.0  & 38.4  & 39.0  & - \\
		BAM \cite{Lang_2022_CVPR}(CVPR’22) &  $641 \times 641$ &       & 43.4  & 50.6  & 47.5  & 43.4  & 46.2  & -  & 49.3  & 54.2 & 51.6  & 49.6  & 51.2  & - \\
		MIANet \cite{yang2023mianet}(CVPR’23)&  $473 \times 473$  &       & 42.5  & 53.0  & 47.8  & 47.4  & 47.7  & 71.5  & 45.8  & 58.2  & 51.3  & 51.9  & 51.7  & 73.1  \\
		SCCAN \cite{xu2023self}(ICCV’23) &  $473 \times 473$ &       & 40.4  & 49.7  & 49.6  & 45.6  & 46.3  & 69.9  & 47.2  & 57.2  & 59.2  & 52.1  & 53.9  & 74.2  \\
		MSI \cite{moon2023msi}(ICCV’23) &  $417 \times 417$ &       & 42.4  & 49.2  & 49.4  & 46.1  & 46.8  & -   & 47.1  & 54.9  & 54.1  & 51.9  & 52.0  & - \\
		AENet \cite{xu2024eliminating}(ECCV’24)&  $473 \times 473$  &      & 43.1 &56.0 &50.3 &48.4 &49.4&73.6  &51.7 &61.9 &\textbf{57.9} &\textbf{55.3} &56.7 &76.5 \\
		ABCB \cite{zhu2024addressing}(CVPR’24)&  $641 \times 641$  &   &44.2 &54.0 &52.1 &49.8 &50.0 &- &50.5 &59.1 &57.0 &53.6 &55.1 &-\\
		
		\cmidrule{1-2}\cmidrule{4-15}    
		OCNet(ours) &  $641 \times 641$ & &  \textbf{45.9} 	&  \textbf{56.9} 	&  \textbf{52.9} 	&  \textbf{50.4} 	&  \textbf{51.5} & \textbf{73.7} & \textbf{52.7}  & \textbf{63.1}  & 57.4 & 54.8 & \textbf{57.0} & \textbf{76.8}  \\

		\bottomrule
	\end{tabular}%
	\vspace{-6pt}
	\label{tab2}%
\end{table*}%

\begin{figure*}[!t]
	\centering
	\centering
	{\scriptsize{\rotatebox{90}{\thinspace}}}\thinspace
	\begin{minipage}{0.107\linewidth}	
		\centering
		{\scriptsize{\thinspace\thinspace Support Image}}
	\end{minipage}
	\begin{minipage}{0.103\linewidth}	
		\centering
		{\scriptsize{\thinspace\thinspace Support Mask}}
	\end{minipage}
	\begin{minipage}{0.104\linewidth}	
		\centering
		{\scriptsize{\thinspace Query Image}}
	\end{minipage}
	\begin{minipage}{0.105\linewidth}	
		\centering
		{\scriptsize{Query Mask}}
	\end{minipage}
	\begin{minipage}{0.105\linewidth}	
		\centering
		{\scriptsize{Baseline}}
	\end{minipage}
	\begin{minipage}{0.103\linewidth}	
		\centering
		{\scriptsize{BAM}}
	\end{minipage}
	\begin{minipage}{0.103\linewidth}	
		\centering
		{\scriptsize\scriptsize{${M}_{g}$}}
	\end{minipage}	
	\begin{minipage}{0.103\linewidth}	
		\centering
		{\scriptsize\scriptsize{${\hat{M}}_{g}$}}
	\end{minipage}	
	\begin{minipage}{0.103\linewidth}	
		\centering
		{\scriptsize{Ours}}
	\end{minipage}\\
	\vspace{0.2ex}
	{\scriptsize{\rotatebox{90}{\qquad Bicycle}}}
	{\includegraphics[width=0.10\textwidth,height=0.10\textwidth]{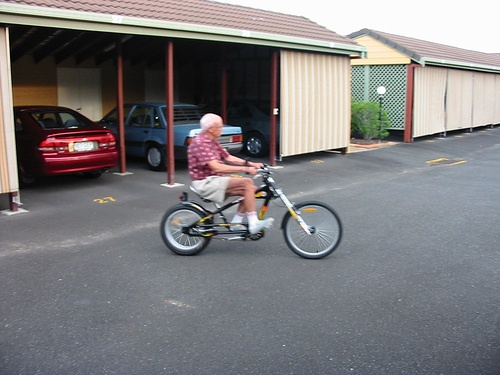}}
	\thinspace
	{\includegraphics[width=0.10\textwidth,height=0.10\textwidth]{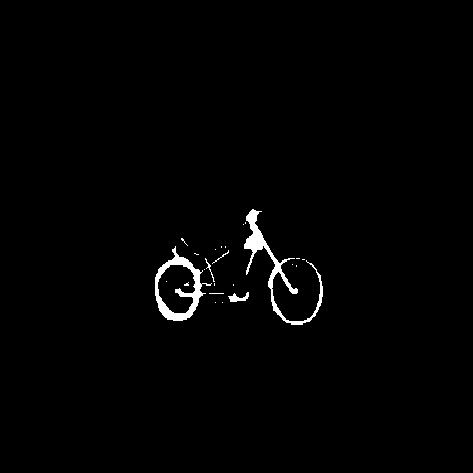}}
	\thinspace
	{\includegraphics[width=0.10\textwidth,height=0.10\textwidth]{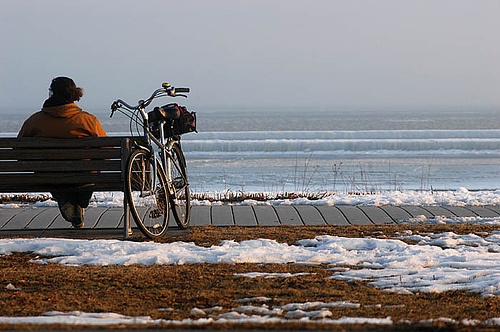}}
	\thinspace
	{\includegraphics[width=0.10\textwidth,height=0.10\textwidth]{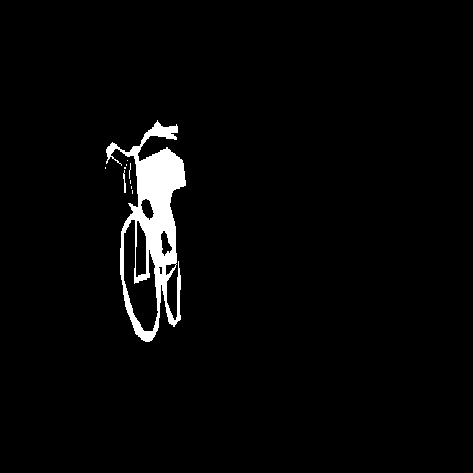}}
	\thinspace
	{\includegraphics[width=0.10\textwidth,height=0.10\textwidth]{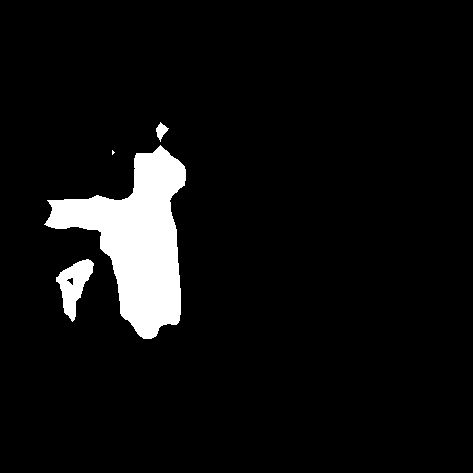}}
	\thinspace
	{\includegraphics[width=0.10\textwidth,height=0.10\textwidth]{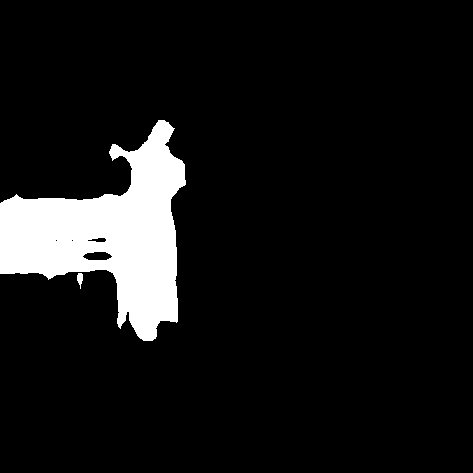}}
	\thinspace
	{\includegraphics[width=0.10\textwidth,height=0.10\textwidth]{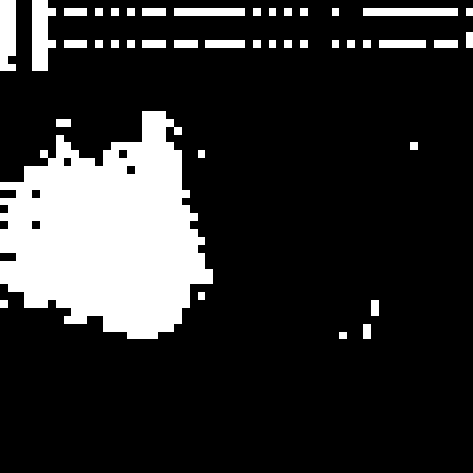}}
	\thinspace
	{\includegraphics[width=0.10\textwidth,height=0.10\textwidth]{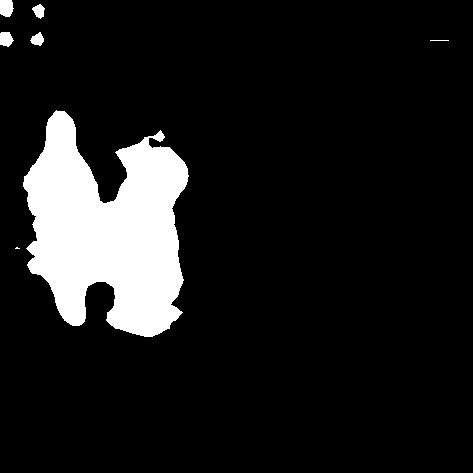}}
	\thinspace
	{\includegraphics[width=0.10\textwidth,height=0.10\textwidth]{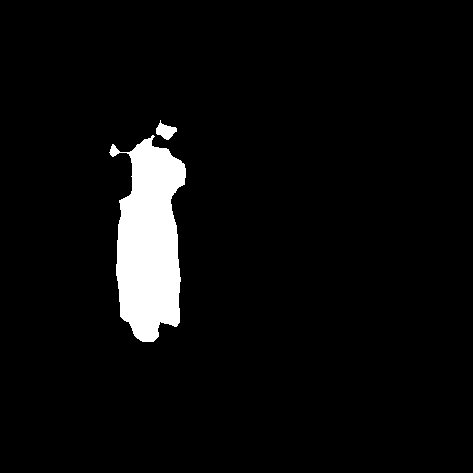}}
	\thinspace
	\vspace{0.5ex}
	
	{\scriptsize{\rotatebox{90}{\qquad Person}}}
	{\includegraphics[width=0.10\textwidth,height=0.10\textwidth]{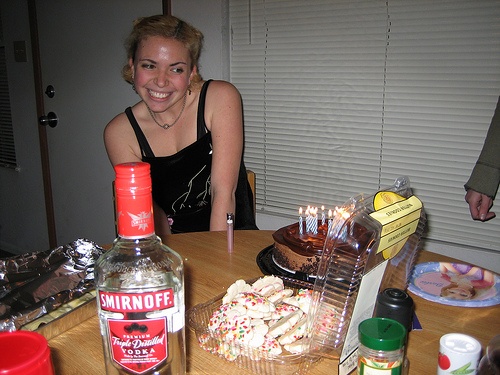}}
	\thinspace
	{\includegraphics[width=0.10\textwidth,height=0.10\textwidth]{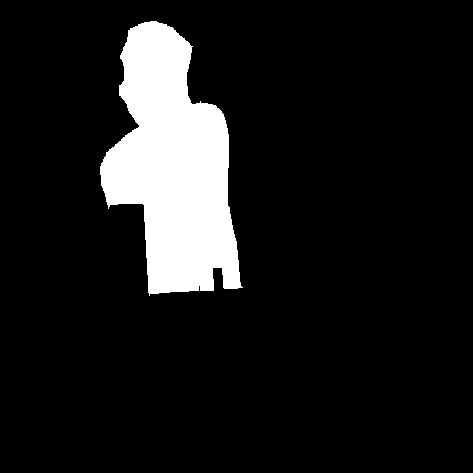}}
	\thinspace
	{\includegraphics[width=0.10\textwidth,height=0.10\textwidth]{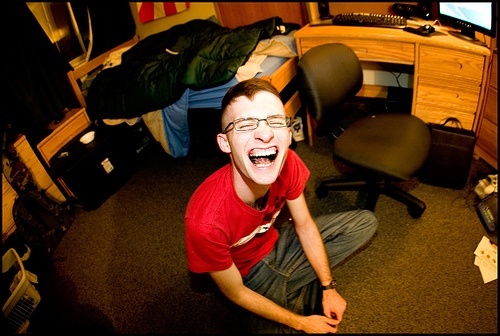}}
	\thinspace
	{\includegraphics[width=0.10\textwidth,height=0.10\textwidth]{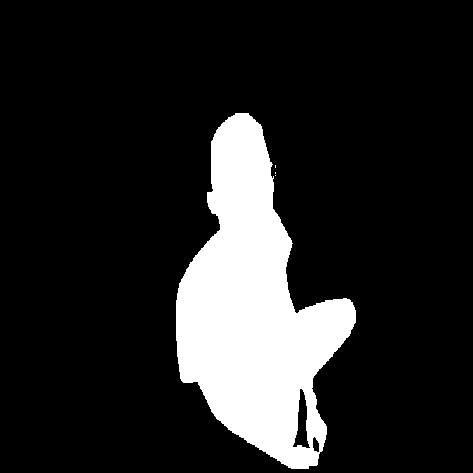}}
	\thinspace
	{\includegraphics[width=0.10\textwidth,height=0.10\textwidth]{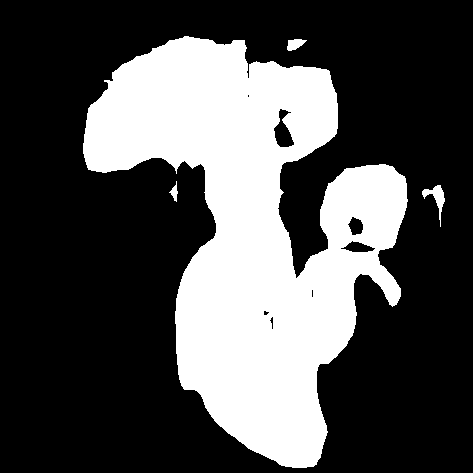}}
	\thinspace
	{\includegraphics[width=0.10\textwidth,height=0.10\textwidth]{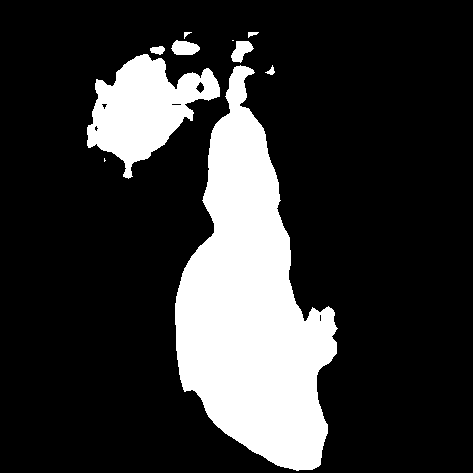}}
	\thinspace
	{\includegraphics[width=0.10\textwidth,height=0.10\textwidth]{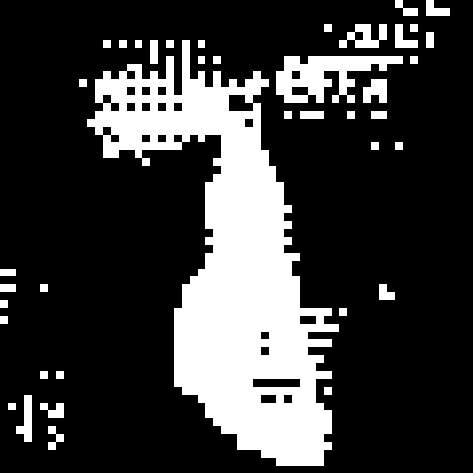}}
	\thinspace
	{\includegraphics[width=0.10\textwidth,height=0.10\textwidth]{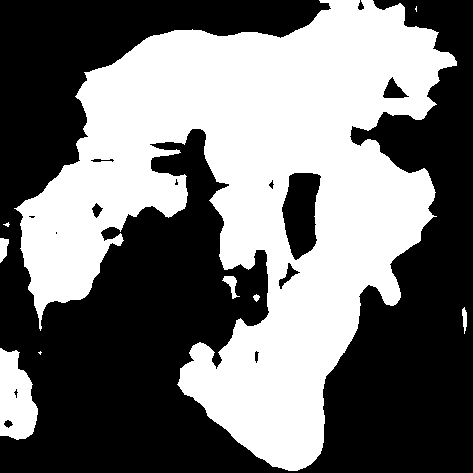}}
	\thinspace
	{\includegraphics[width=0.10\textwidth,height=0.10\textwidth]{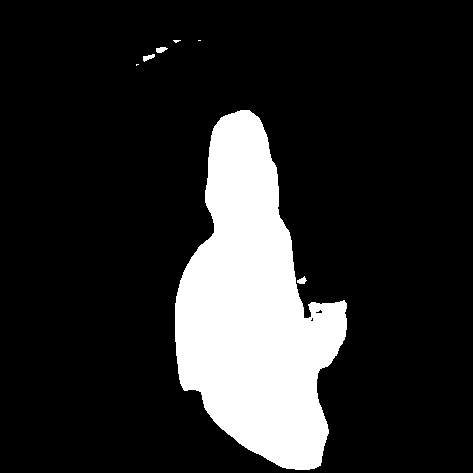}}
	\thinspace
	
	\vspace{0.5ex}
	{\scriptsize{\rotatebox{90}{\qquad\quad Car}}}
	{\includegraphics[width=0.10\textwidth,height=0.10\textwidth]{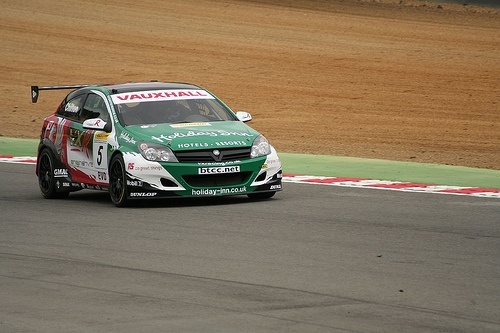}}
	\thinspace
	{\includegraphics[width=0.10\textwidth,height=0.10\textwidth]{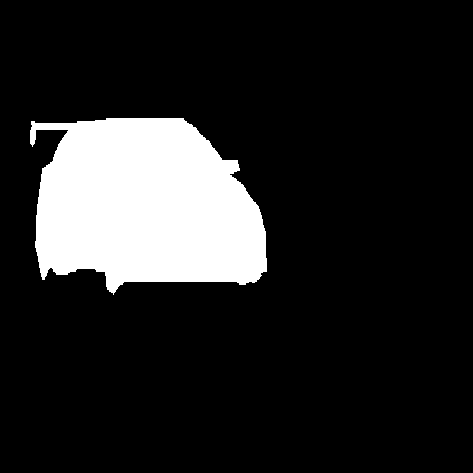}}
	\thinspace
	{\includegraphics[width=0.10\textwidth,height=0.10\textwidth]{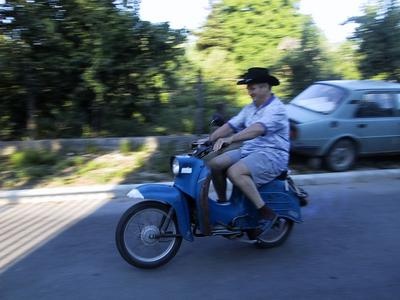}}
	\thinspace
	{\includegraphics[width=0.10\textwidth,height=0.10\textwidth]{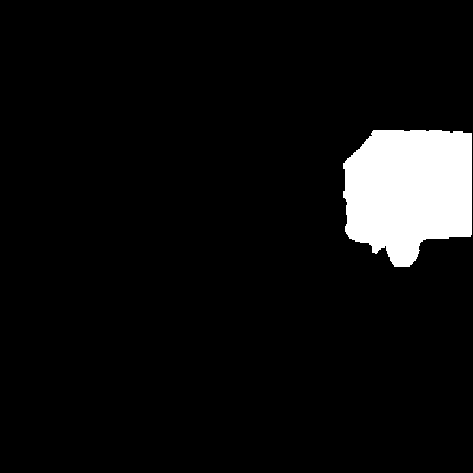}}
	\thinspace
	{\includegraphics[width=0.10\textwidth,height=0.10\textwidth]{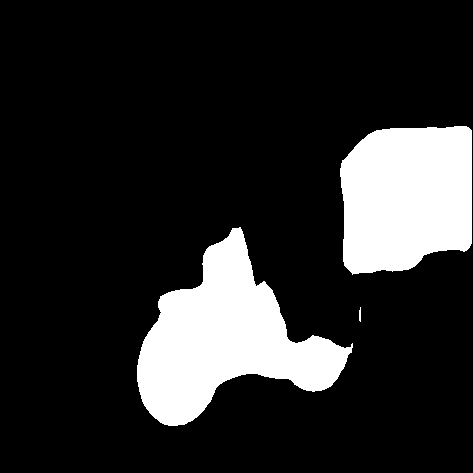}}
	\thinspace
	{\includegraphics[width=0.10\textwidth,height=0.10\textwidth]{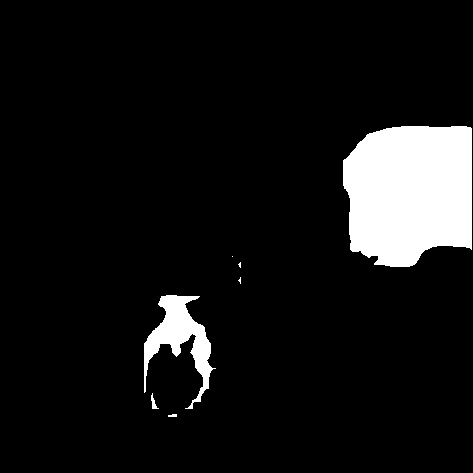}}
	\thinspace
	{\includegraphics[width=0.10\textwidth,height=0.10\textwidth]{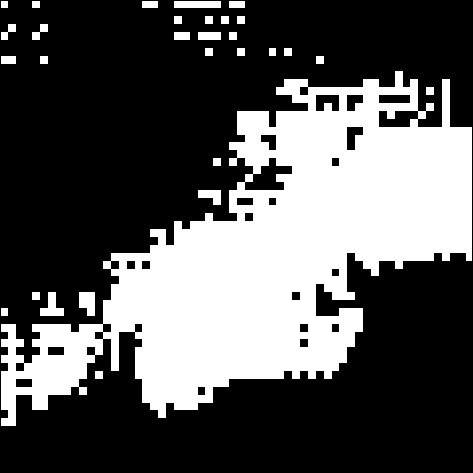}}
	\thinspace
	{\includegraphics[width=0.10\textwidth,height=0.10\textwidth]{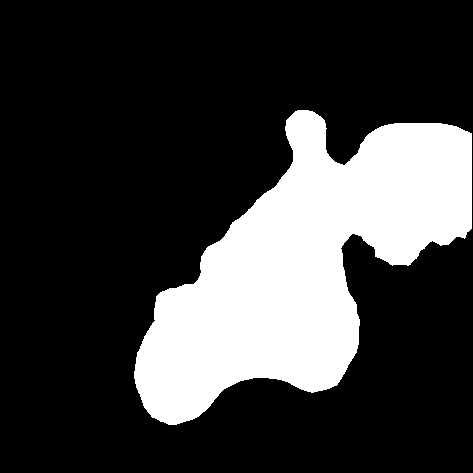}}
	\thinspace
	{\includegraphics[width=0.10\textwidth,height=0.10\textwidth]{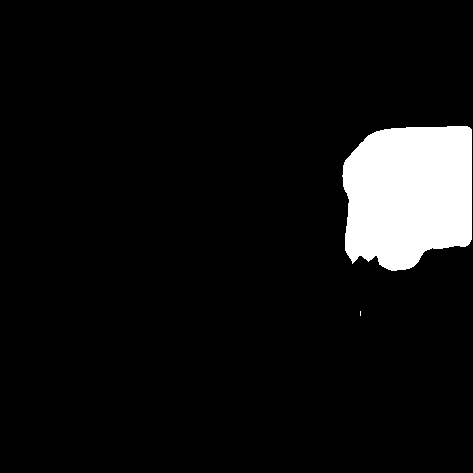}}
	\thinspace
	
	\vspace{0.5ex}
	{\scriptsize{\rotatebox{90}{\qquad\enspace Horse}}}
	{\includegraphics[width=0.10\textwidth,height=0.10\textwidth]{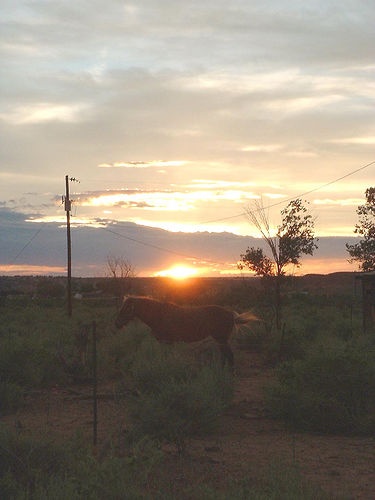}}
	\thinspace
	{\includegraphics[width=0.10\textwidth,height=0.10\textwidth]{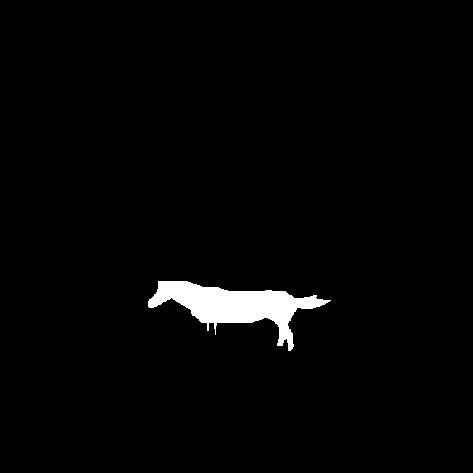}}
	\thinspace
	{\includegraphics[width=0.10\textwidth,height=0.10\textwidth]{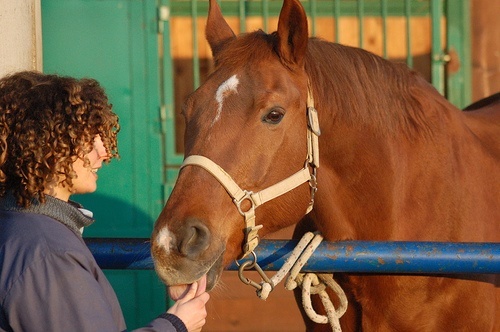}}
	\thinspace
	{\includegraphics[width=0.10\textwidth,height=0.10\textwidth]{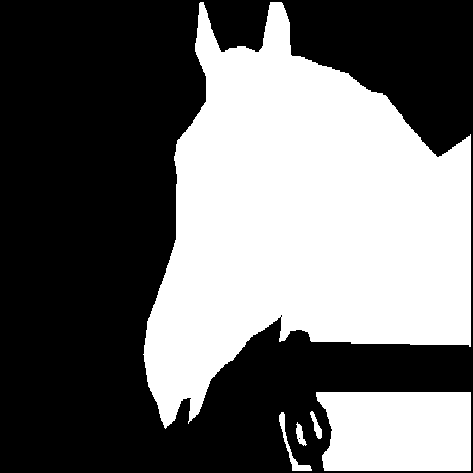}}
	\thinspace
	{\includegraphics[width=0.10\textwidth,height=0.10\textwidth]{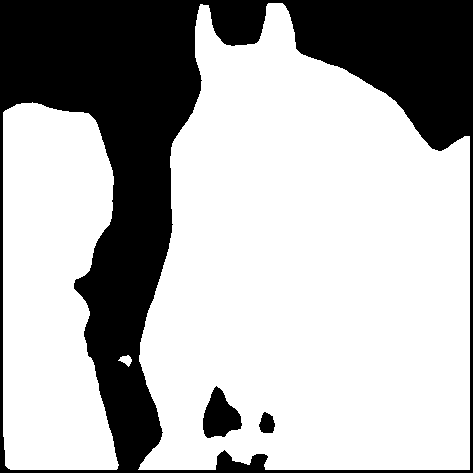}}
	\thinspace
	{\includegraphics[width=0.10\textwidth,height=0.10\textwidth]{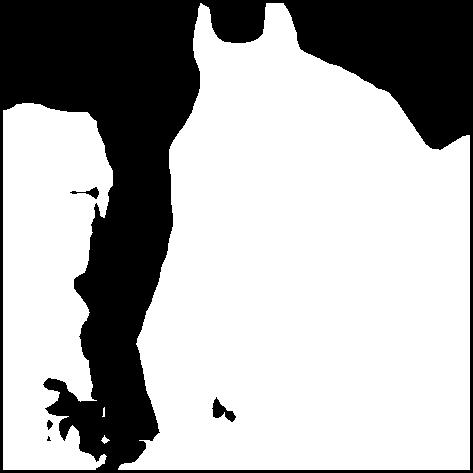}}
	\thinspace
	{\includegraphics[width=0.10\textwidth,height=0.10\textwidth]{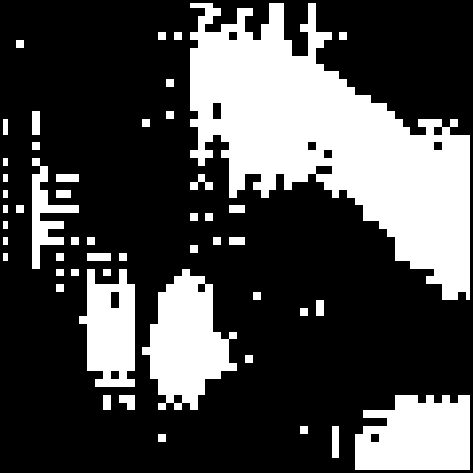}}
	\thinspace
	{\includegraphics[width=0.10\textwidth,height=0.10\textwidth]{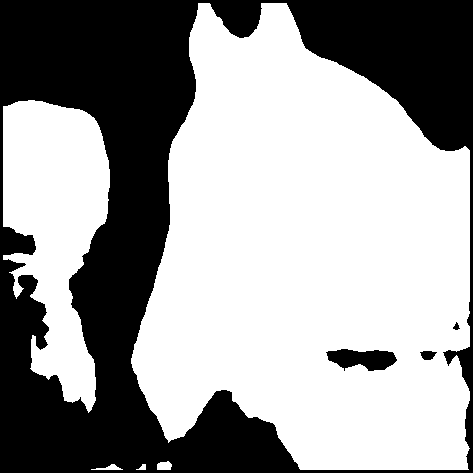}}
	\thinspace
	{\includegraphics[width=0.10\textwidth,height=0.10\textwidth]{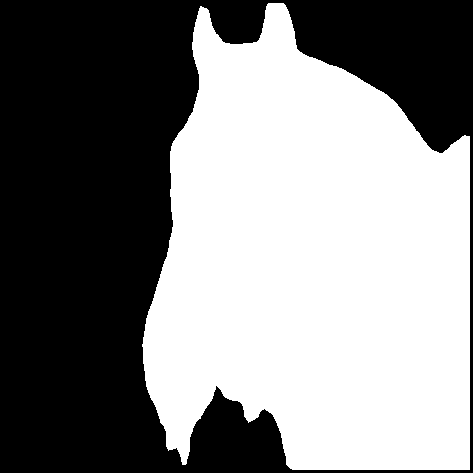}}
	\thinspace
	\vspace{-5pt}	
	\caption{Qualitative results of our OCNet, baseline method, BAM and general objects. Each column from left to right represents the support image, support mask, query image, query mask, baseline prediction, BAM \cite{Lang_2022_CVPR} prediction, general object mask ${M}_{g}$, general object prediction ${\hat{M}}_{g}$ and our OCNet prediction, respectively.}
	\label{fig4}
	\vspace{-16pt}
\end{figure*}

\subsection{Comparison with State-of-the-Arts}
\vspace{-5pt}
\textbf{Quantitative results.}\hspace{0.5em}
To evaluate the effectiveness of OCNet, we present the quantitative comparisons between it and other state-of-the-art FSS approaches in Table \ref{tab1} and Table \ref{tab2}. It can be observed that the proposed OCNet outperforms previous advanced approaches on PASCAL-$5^{i}$ and COCO-$20^{i}$ datasets under all settings. Specifically, with the VGG-16 backbone, the proposed method achieves 0.9\% (1-shot) and 0.8\% (5-shot) of mIoU improvements over the previous best results on PASCAL-$5^{i}$. Moreover, OCNet also achieves nearly 1\% improvement with ResNet-50 as the backbone. Therefore, compared to other image-level correlation methods with post-elimination processing (\textit{i.e.}, BAM~\cite{Lang_2022_CVPR}, ABCB \cite{zhu2024addressing} and so on), our object-level correlation method is more effective for mining the target object and suppressing the hard pixel noise. As for more challenging COCO-$20^{i}$, the performances reach 51.5\% (1-shot) and 57.0\% (5-shot) of mIoU with ResNet-50 backbone, surpassing previous state-of-the-art in Table \ref{tab2}. Besides, the FB-IoU results also achieve improvements, especially for the 1-shot results on the PASCAL-$5^{i}$. \vspace{3pt}
\\\textbf{Qualitative results.}\hspace{0.5em}
To better analyze and understand the proposed OCNet, we report some qualitative results generated from our OCNet, baseline model, and BAM \cite{Lang_2022_CVPR} in Fig. \ref{fig4}. The baseline model is established by removing the GOMM and CCM. From Fig. \ref{fig4}, we can observe that BAM and baseline model always falsely activate the irrelevant background objects with the image-level correlation in spite of the post-elimination, such as the person and chair in the first row, the bed and chair in the second row, the motorcycle and person in the third row, and the person in the last row. Different from the BAM and baseline model, this irrelevant information can be mined by the general object prototypes in our model, as shown in the general object prediction ${\hat{M}}_{g}$. Then, by further establishing the object-level correlation between the support target feature and the query general object feature, our OCNet accurately identifies the target object while suppressing the irrelevant objects. Moreover, by comparing the results of ${M}_{g}$ and ${\hat{M}}_{g}$, it can demonstrate that the moderately uncompleted information favors the generalization and reconstruction ability of general object prototypes.
\subsection{Ablation Study}
\hspace{1em}We conduct the ablation study with ResNet-50 backbone under the 1-shot setting on the PASCAL-$5^{i}$ dataset. 
\begin{table}
	\belowrulesep=0pt
	\aboverulesep=0pt
	\centering \scriptsize
	\setlength{\tabcolsep}{4pt}
	\renewcommand{\arraystretch}{1.2}
	\caption{Ablation studies of main components in OCNet.}
	\vspace{-5pt}
	\begin{tabular}{c|c|cccc|c}
		\toprule
		GOMM  & CCM  & \multicolumn{1}{c}{Fold0} & \multicolumn{1}{c}{Fold1} & \multicolumn{1}{c}{Fold2} & \multicolumn{1}{c|}{Fold3} & \multicolumn{1}{c}{Mean} \\
		\midrule
		\multicolumn{1}{c|}{} & \multicolumn{1}{c|}{} & 67.5		&73.4		&66.5		&61.6		&67.3 \\
		\checkmark     & \multicolumn{1}{c|}{} &69.9	&	74.2	&	68.3		&63.9	&	69.1\\
		\multicolumn{1}{c|}{} & \checkmark     &71.9		&74.7		&69.8		&63.0		&69.9 \\
		\checkmark     & \checkmark     &  \textbf{73.5} & \textbf{75.9} & \textbf{71.1} & \textbf{64.9}  & \textbf{71.4} \\
		\bottomrule
	\end{tabular}%
	\vspace{-5pt}
	\label{tab3}%
\end{table}%
\\\textbf{Effect of GOMM and CCM.}\hspace{0.5em}
Table \ref{tab3} reports the ablation results regarding the effectiveness of the proposed General Object Mining Module (GOMM) and Correlation Construction Module (CCM). The first line denotes the baseline result (67.3\%), where the baseline is established by the feature extractor and the decoder. As shown in Table \ref{tab3}, when integrating the GOMM into the baseline, class mIoU significantly increases by 1.8\%. It proves that it is more effective to establish the correlation with the query general object feature than the entire image feature. Besides, CCM exploits richer support information and models a more suitable allocation pattern compared to the dense global prototype comparison used in the baseline, resulting in a 2.6\% improvement in mIoU (69.9\% vs. 67.3\%) .Finally, when both GOMM and CCM are employed, the performance further improves to 71.4\% (an improvement of 4.1\%), demonstrating the effectiveness of each module and the benefit of establishing object-level correlations.
\begin{table}
	\belowrulesep=0pt
	\aboverulesep=0pt
	\centering \scriptsize
	\setlength{\tabcolsep}{4pt}
	\renewcommand{\arraystretch}{1.2}
	\caption{Ablation studies on general object mask in GOMM.}
	\vspace{-5pt}
	\begin{tabular}{l|cccc|c}
		\toprule
		\multicolumn{1}{r|}{} & \multicolumn{1}{c}{Fold0} & \multicolumn{1}{c}{Fold1} & \multicolumn{1}{c}{Fold2} & \multicolumn{1}{c|}{Fold3} & \multicolumn{1}{c}{Mean} \\
		\midrule
		&71.9		&74.7		&69.8		&63.0		&69.9 \\
		CAM & 71.5	&75.4		&70.6		&63.4		&70.3 \\
		CAM + Mask &72.4		&75.5		&70.9		&64.1		&70.7 \\
		CAM + Atten &72.5		&75.4		&71.0		&64.5		&70.9 \\
		CAM + Mask + Atten & \textbf{73.5} & \textbf{75.9} & \textbf{71.1} & \textbf{64.9}  & \textbf{71.4} \\
		\bottomrule
	\end{tabular}%
	\vspace{-12pt}
	\label{tab4}%
\end{table}%
\\\textbf{General Object Feature in GOMM.}\hspace{0.5em} The general object information generated by the GOMM is learned from three components: CAM, high-level cosine similarity mask, and cross-attention. Table \ref{tab4} presents our validation experiment on the effectiveness of each component, where the first line denotes only using CCM. As shown in the table, when we integrate CAM to mine the general object information from query images, the performance improves from 69.9\% to 70.3\%. This demonstrates that the object-level correlation is effective, even using the incomplete and obscure object information. Then, the similarity mask and the cross-attention are utilized to complement the object information lost by the CAM, contributing to 0.4\% and 0.6\% performance gain, respectively. Moreover, by integrating all the components, we can obtain another 1.5\% performance gain and improve the result to 71.4\%. Obviously, through learning the general object prototypes from three components, GOMM effectively mines rich and complete general object information to build the object-level correlation.
\begin{table}
	\belowrulesep=0pt
	\aboverulesep=0pt
	\centering \scriptsize
	\setlength{\tabcolsep}{4pt}
	\renewcommand{\arraystretch}{1.2}
	\caption{Ablation studies on different allocation methods in CCM.}
	\vspace{-5pt}
	\begin{tabular}{l|cccc|c}
		\toprule
		\multicolumn{1}{c|}{} & \multicolumn{1}{c}{Fold0} & \multicolumn{1}{c}{Fold1} & \multicolumn{1}{c}{Fold2} & \multicolumn{1}{c|}{Fold3} & \multicolumn{1}{c}{Mean} \\
		\midrule
		&69.9		&74.2		&68.3		&63.9		&69.1 \\
		Mean &71.7		&75.5		&69.6		&64.4		&70.3	 \\
		Cosine &72.0	&75.5		&70.5		&64.3		&70.6 \\
		Fore & 72.5		&75.5		&70.0		&64.1		&70.5	 \\
		Fore + Back & \textbf{73.5} & \textbf{75.9} & \textbf{71.1} & \textbf{64.9}  & \textbf{71.4} \\
		\bottomrule
	\end{tabular}%
	\vspace{-12pt}
	\label{tab5}%
\end{table}%
\\\textbf{Prototype Allocation in CCM.}\hspace{0.5em} We apply several prototype allocating methods in CCM and compare their performance in Table \ref{tab5} to explore their effectiveness in correlation construction. The first line denotes only using GOMM. The Mean denotes concatenating the global prototype and the average of the remaining selected prototypes without any allocation. In Cosine, the prototypes are allocated by the cosine similarity. Fore and Back mean the foreground and background prototypes are allocated by the supervision of prototype allocating mask ${M}_{pa}$. Compared with other methods (Mean and Cosine), our proposed ${M}_{pa}$ (Fore+Back) delivers the optimal allocation method for the prototype, obtaining 71.4\% performance. Moreover, by comparing the results from Fore and Fore+Back, we argue that the foreground and background prototypes are complementary and necessary in correlation construction.

	\vspace{-8pt}
\section{Conclusion}
\vspace{-3pt}
\label{sec:conclusion}
\hspace{1em}In this paper, we propose a novel Object-level Correlation Network (OCNet) to model the the query-support correlation in the FSS from a new perspective. Instead of the previous image-level correlation methods, our OCNet establishes the object-level correlation between the support target objects and the query general objects by mimicking the process of biological vision. Target identification in the general objects is more valid than in the entire image, especially in the low-data regime. In this way, the query target object can be accurately identified while the other irrelevant objects are suppressed. Extensive experimental results demonstrate the superiority of OCNet. In the future, we believe that it is a promising direction to explore further possibilities of object-level correlations in other few-shot scenarios.
\\\textbf{Acknowledgements}\hspace{0.5em} This research work is supported by National Key R\&D Program of China (2021YFA1001100), National Natural Science Foundation of China under Grant (62272231), CIE-Tencent Robotics X Rhino-Bird Focused Research Program,  the Fundamental Research Funds for the Central Universities (4009002401), the EDB Space Technology Development Programme under Project S22-19016-STDP, and  the Big Data Computing Center of Southeast University.
	{\small
		\bibliographystyle{ieeenat_fullname}
		\bibliography{11_references}
	}

	
\end{document}